\documentclass[letterpaper]{article} % DO NOT CHANGE THIS
\usepackage[preprint]{aaai2027}  % Public preprint: suppress AAAI publication footer.
\usepackage[hyphens]{url}  % DO NOT CHANGE THIS
\usepackage{graphicx} % DO NOT CHANGE THIS
\graphicspath{{./}} % Keep figure paths relative to the arXiv source root.
\usepackage{natbib}  % DO NOT CHANGE THIS AND DO NOT ADD ANY OPTIONS TO IT
\usepackage{caption} % DO NOT CHANGE THIS AND DO NOT ADD ANY OPTIONS TO IT
\usepackage{amsmath}
\usepackage{amssymb}
\usepackage{booktabs}
\usepackage{multirow}

\newcommand{\method}{SemiMat}
\newcommand{\matrank}{MatRank}

\title{Learning Materials Properties from Scarce Labels and Unlabeled Crystals}
\author{
Wentao Li\textsuperscript{\rm 1,$\dagger$},
Yizhe Chen\textsuperscript{\rm 1,$\dagger$},
Jiangjie Qiu\textsuperscript{\rm 1},
Yijun Li\textsuperscript{\rm 1},
Leyi Zhao\textsuperscript{\rm 1},
Xiaonan Wang\textsuperscript{\rm 1,*}
}
\affiliations{
\textsuperscript{\rm 1}Beijing Key Laboratory of Artificial Intelligence for
Advanced Chemical Engineering Materials,\\
State Key Laboratory of Chemical Engineering and Low-Carbon Technology,\\
Department of Chemical Engineering, Tsinghua University\\
\textsuperscript{$\dagger$}These authors contributed equally.\quad
\textsuperscript{*}Corresponding author.
}

\begin{document}

\maketitle

\begin{abstract}
Learning materials properties from scarce labels and unlabeled crystals is a
central challenge for data-driven materials discovery. We present \method, a
controlled benchmark for semi-supervised materials property regression, and
\matrank, a reliability-weighted objective for continuous pseudo-label
uncertainty. \method\ fixes labeled and unlabeled crystal inputs, graph-backbone
interfaces, validation-only checkpoint selection, held-out test reporting,
normalized MAE (NMAE), and method-rank summaries across six scarce-label tasks,
four graph backbones, and five predefined split runs. \matrank\ builds pseudo-targets from
labeled anchors, weights them by local reliability and weak-prediction
agreement, trains weak and strong graph views consistently, and adds ranking
signals so that unlabeled crystals shape both values and candidate order. Across
the retained 24 backbone-task blocks, one fixed \matrank\ objective gives the
lowest aggregate held-out test NMAE ($0.896$) and best average method rank
($2.208$). The component, OOD, and generated-pool diagnostics identify where
the gain is reliable and where further screening evaluation remains necessary.
Code is available at \url{https://github.com/littlepeachs/SemiMat}.
\end{abstract}

\section{Introduction}
Materials discovery increasingly requires learning useful property predictors
before enough reliable labels exist. Public repositories, high-throughput
calculations, and large-scale materials models have expanded the space of
candidate crystals, but experimental measurements and high-fidelity simulations
remain expensive, uneven, and property dependent
\cite{butler2018machine,merchant2023scaling,jain2013materials,choudhary2020jarvis}.
The bottleneck is therefore not only how accurately a model predicts after
labels are available. It is whether learning systems can use abundant unlabeled
structures without converting their own uncertain predictions into misleading
supervision.

This setting is a difficult form of semi-supervised regression. In
classification, confident pseudo-labels can often be thresholded and consistency
regularization can be tied to discrete decisions \cite{chapelle2006ssl}. In
materials property prediction, the target is continuous, task scales differ by
orders of magnitude, and a numerically sharp pseudo-label can still be wrong.
The unlabeled pool is also part of the scientific question: a pool sampled from
a broad materials database, a shifted composition range, or a generative model
may help representation learning in one task while injecting misleading
structure in another.

The field also lacks a controlled protocol for deciding when such unlabeled
signals genuinely help. Apparent gains can depend on the property, graph
backbone, train/validation/test split, unlabeled pool, seed, and aggregation
rule. A single aggregate score hides whether a method improves scarce-label
regression broadly, succeeds only with a favorable encoder, or benefits from a
particular unlabeled source. Materials screening adds another constraint:
decisions depend on continuous values, yet the order in which candidates are
inspected often determines which structures receive further computation or
experimental attention.

\begin{figure*}[t]
\centering
\includegraphics[width=0.96\textwidth]{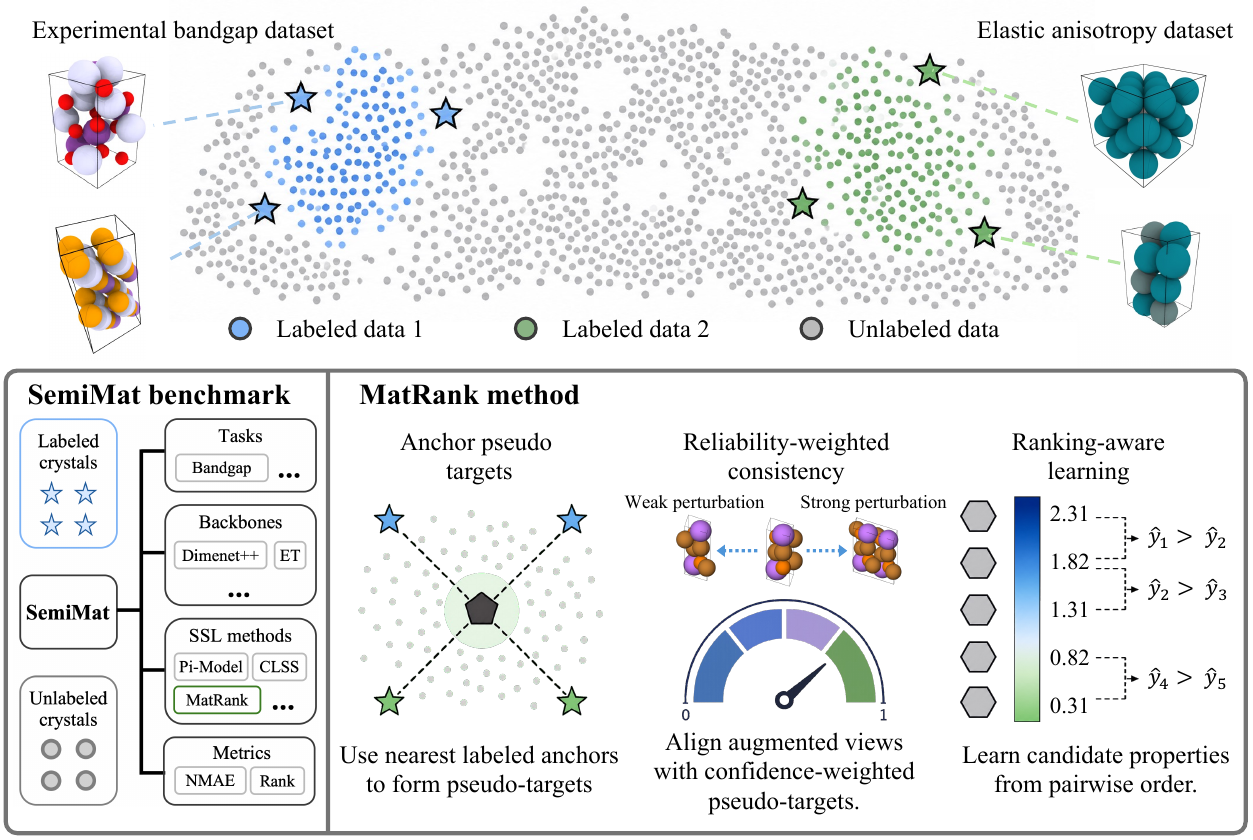}
\caption{SemiMat benchmark and MatRank training design. \method\ fixes labeled
and unlabeled inputs, tasks, graph backbones, semi-supervised methods and
metrics for auditable scarce-label regression. \matrank\ turns unlabeled
crystals into reliability-weighted supervision through anchor pseudo-targets,
weak--strong view alignment and pairwise order constraints. The workflow
separates the benchmark contract from the learning objective while tying both
to held-out test evaluation.}
\label{fig:overview}
\end{figure*}

We address this problem with \method, a controlled benchmark and framework for
semi-supervised materials property regression, together with \matrank, a
reliability-weighted algorithm for continuous pseudo-label uncertainty. \method\
fixes labeled and unlabeled inputs, graph-backbone interfaces, validation-only
checkpoint selection, held-out test reporting, split-level MAE, normalized MAE
(NMAE), and method-rank summaries across six scarce-label materials tasks, four
graph backbones, and five predefined split runs. \matrank\ builds pseudo-targets from labeled
anchors, weights them by local reliability and weak-prediction agreement, trains
weak and strong graph views consistently, and adds ranking signals so that
unlabeled structures shape both values and candidate order. Across the retained
24 backbone-task blocks, one fixed \matrank\ objective gives the lowest
aggregate held-out test NMAE and best average method rank under
validation-selected checkpoints, while the ablation, OOD, and generated-pool
experiments define the current evidence boundary. Figure~\ref{fig:overview}
summarizes the benchmark contract and the \matrank\ training signals.

\section{Related Work}
Machine learning has become a central tool for molecular and materials
discovery \cite{butler2018machine,merchant2023scaling}. Large materials
repositories and benchmarks have improved data access and model comparison
\cite{jain2013materials,choudhary2020jarvis,dunn2020matbench}, but label
coverage remains uneven across properties. Experimental measurements,
stability-related targets, tensorial responses, and expensive high-fidelity
labels are often much smaller than the pool of candidate structures. \method\
targets this mismatch by treating unlabeled structures as part of the learning
protocol, not as unused background data.

Crystal graph learning provides the backbone layer for this protocol. These
models build on message passing and graph convolution ideas
\cite{gilmer2017neural,kipf2017semi} and, for 3D structures, increasingly use
geometric equivariance \cite{satorras2021egnn,batzner2022nequip}. CGCNN, MEGNet,
SchNet, DimeNet++, GemNet, ALIGNN, M3GNet, and equivariant transformer-style
models encode atomistic geometry with different inductive biases
\cite{xie2018cgcnn,chen2019megnet,schutt2018schnet,klicpera2020dimenet,
gasteiger2020dimenetpp,gasteiger2021gemnet,choudhary2021alignn,chen2022m3gnet,
tholke2022torchmdnet,liao2023equiformer}. \method\ does not introduce a new
encoder; it asks whether semi-supervised objectives improve regression when the
backbone is controlled.

Semi-supervised learning commonly uses entropy minimization, pseudo-labeling,
consistency regularization, teacher-student targets, and augmentation-based
label guessing \cite{chapelle2006ssl,grandvalet2004entropy,lee2013pseudo,
laine2017temporal,tarvainen2017mean,berthelot2019mixmatch,xie2020uda,
sohn2020fixmatch}. Realistic SSL studies also show that unlabeled data can hurt
when the protocol or pool is mismatched \cite{oliver2018realistic}. For
materials regression, this risk is amplified because confidence is harder to
calibrate for continuous pseudo-labels than for classes. \matrank\ therefore
uses pairwise learning-to-rank supervision \cite{burges2005ranknet} as a
training signal that complements pointwise regression.

\section{Problem Setup}
Let $D_l=\{(G_i,y_i)\}_{i=1}^{n}$ be labeled materials graphs with scalar
properties $y_i\in\mathbb{R}$, and let $D_u=\{U_j\}_{j=1}^{m}$ be an unlabeled
candidate pool available during training. \method\ fixes the task, split,
backbone, seed, and unlabeled pool before comparing algorithms. The question is
whether unlabeled structures improve scarce-label regression without using test
labels for training, checkpoint selection, or hyperparameter tuning. A graph
encoder $g_\theta$ and regression head $r_\theta$ predict
\begin{equation}
  \hat{y}_i = r_\theta(g_\theta(G_i)).
\end{equation}
Reported MAE is computed after returning predictions to the original property
scale:
\begin{equation}
  \mathrm{MAE}=\frac{1}{|D|}\sum_{(G,y)\in D}|\hat{y}-y|.
\end{equation}

Because tasks have different units and scales, raw MAE cannot be averaged
directly across datasets. We compute the five-split mean MAE for every method
within a fixed backbone-task block and normalize it by the supervised mean MAE
in that same block:
\begin{equation}
  \mathrm{NMAE}_{m,b,t} =
  \frac{\mathrm{MAE}_{m,b,t}}
       {\mathrm{MAE}_{\mathrm{Supervised},b,t}}.
\end{equation}
Lower NMAE is better; supervised training is therefore exactly $1.0$ in every
block. We also compute method rank within each backbone-task block, which
summarizes how often an algorithm appears near the top of local MAE rankings.

\section{SemiMat Framework}
\method\ specifies a benchmark contract before training: the labeled
train/validation/test split, unlabeled pool, graph-construction interface,
backbone, algorithm, seed, and reporting metrics are fixed for each block. The
same validation split selects checkpoints for every method, and the held-out
test split is used only for final reporting. This separates the benchmark
question, whether unlabeled structures help under a controlled protocol, from
the method question, whether a new objective improves that protocol.

The framework has four layers. The data layer exposes matched labeled and
unlabeled structures; the encoder layer supplies SchNet, DimeNet++, ET, or
M3GNet \cite{schutt2018schnet,gasteiger2020dimenetpp,tholke2022torchmdnet,
chen2022m3gnet}; the algorithm layer runs supervised learning, Pi Model, Mean
Teacher, MixMatch, RDA, CLSS, or \matrank\
\cite{laine2017temporal,tarvainen2017mean,berthelot2019mixmatch,
huang2024rankup,dai2023clss}; and the evaluation layer reports split-level MAE,
five-split means, NMAE, and method ranks. Full backbone and baseline descriptions
are provided in the supplement.

The contract also specifies exclusions. The final \matrank\ row is not a
per-dataset selector, a per-backbone selector, or a validation rule that chooses
among several algorithms. It is one objective applied unchanged to each
backbone-task block. This matters for materials discovery because a local improvement
can be misleading when tasks differ in label noise, target scale, and geometric
complexity. The benchmark therefore exposes both aggregate behavior and local
exceptions.

\paragraph{MatRank.}
\matrank\ is the new algorithm within \method. It is motivated by a regression
failure mode: an unlabeled crystal can receive a continuous pseudo-label that is
numerically precise but unreliable. \matrank\ therefore combines four signals:
labeled-anchor pseudo-targets, reliability-weighted weak--strong consistency,
cross-set and labeled-batch ranking, and an auxiliary ranking-classifier (ARC)
head. The method uses one code path and one fixed objective for all datasets,
backbones, and indexed splits.

The algorithm is deliberately conservative about pseudo-targets. A nearby
labeled anchor is useful when the local neighborhood is sharp and the weak
prediction agrees with the anchor estimate. When either condition fails,
\matrank\ reduces the influence of the anchor and relies more on consistency
and order constraints. This design keeps the method aligned with the main
scientific use case: unlabeled structures should help shape the representation,
but they should not inject high-weight numerical targets when the local
evidence is weak.

For an unlabeled weak view $u^w$ with embedding $h_u^w$, weak prediction
$p_u^w=f_\theta(u^w)$, and strong prediction $p_u^s=f_\theta(u^s)$,
\matrank\ first retrieves labeled anchors in the current batch. Let
$s_{ui}=\cos(h_u^w,h_i)$, $\mathcal{N}_K(u)$ be the top-$K$ anchors, and
$\nu(\cdot)$ be the same target normalizer used by supervised training:
\begin{equation}
  \begin{aligned}
  \alpha_{ui} &=
  \frac{\exp(s_{ui}/\tau_a)}
       {\sum_{j\in \mathcal{N}_K(u)} \exp(s_{uj}/\tau_a)},\\
  \hat{y}_u &=
  \nu\!\left(\sum_{i\in \mathcal{N}_K(u)}\alpha_{ui}y_i\right).
  \end{aligned}
\end{equation}
Anchor trust combines local neighborhood sharpness and agreement with the
weak-view regressor. Let $\Delta s_u=s_{u,(1)}-s_{u,(K)}$ and
$I_b=[\ell_{\min}-m_c,\ell_{\max}+m_c]$:
\begin{equation}
  \begin{aligned}
  \rho_u &=
  \mathrm{clip}\!\left[
  \sigma\!\left(\gamma_A\Delta s_u\right)
  \exp\!\left(-q|p_u^w-\hat{y}_u|/\gamma_P\right),0,1\right],\\
  z_u &=
  \Pi_{I_b}\!\left(\rho_u\hat{y}_u+(1-\rho_u)\,\mathrm{sg}(p_u^w)\right).
  \end{aligned}
\end{equation}
Here $\ell_{\min}$ and $\ell_{\max}$ are the minimum and maximum normalized
labeled targets in the batch, and $\Pi$ denotes projection onto that interval.
High-reliability anchors move the strong prediction toward nearby labeled
values; low-reliability anchors reduce to consistency with the weak prediction.

The ARC head predicts pairwise order from the same graph representation. On
labeled pairs it is trained by the sign of $y_i-y_j$. On unlabeled data, weak
ARC probabilities provide hard pseudo-labels for the strong view when their
confidence exceeds a threshold. The unlabeled ARC weight
\begin{equation}
  \lambda_{\mathrm{arc}}^u =
  \mathrm{clip}(\eta_0+\eta_1(1-\bar{\rho}),\eta_{\min},\eta_{\max})
\end{equation}
increases when batch-level anchor reliability is low, so order consistency can
carry more of the unlabeled signal when continuous pseudo-targets are fragile.

The regression branch then uses both absolute consistency and relative order.
Let $H_\beta$ denote SmoothL1 loss and
\[
  R_\delta(a,b,\hat a,\hat b)
  =\mathbf{1}\{|a-b|>\delta\}
  [m_r-\operatorname{sgn}(a-b)(\hat a-\hat b)]_+ .
\]
With $\langle\cdot\rangle$ denoting the corresponding normalized average,
\begin{equation}
  \begin{aligned}
  \mathcal{L}_{\mathrm{cons}} &=
  \left\langle \rho_u H_\beta(p_u^s,z_u)\right\rangle_u,\\
  \mathcal{L}_{\mathrm{cross}} &=
  \left\langle \rho_u
  R_{\delta_{lu}}(\nu(y_i),z_u,p_i,p_u^s)\right\rangle_{i,u},\\
  \mathcal{L}_{\mathrm{label}} &=
  \left\langle
  R_{\delta_{ll}}(\nu(y_i),\nu(y_j),p_i,p_j)\right\rangle_{i,j}.
  \end{aligned}
\end{equation}
The weak-view consistency term has the same form as
$\mathcal{L}_{\mathrm{cons}}$ with $p_u^s$ replaced by $p_u^w$, and
$\mathcal{L}_{\mathrm{smooth}}$ matches weak predictions to their
feature-neighbor average. The full fixed objective is
\begin{equation}
  \begin{aligned}
  \mathcal{L}
  &= \mathcal{L}_{\mathrm{sup}} + \lambda_u \mathcal{L}_{\mathrm{ssl}},\\
  \mathcal{L}_{\mathrm{ssl}}
  &= \mathcal{L}_{\mathrm{arc}} +
  \mathcal{L}_{\mathrm{cons}} +
  \omega_{\mathrm{cross}}\mathcal{L}_{\mathrm{cross}} +
  \omega_{\mathrm{weak}}\mathcal{L}_{\mathrm{weak}} \\
  &\quad +
  \omega_{\mathrm{smooth}}\mathcal{L}_{\mathrm{smooth}} +
  \omega_{\mathrm{label}}\mathcal{L}_{\mathrm{label}} .
  \end{aligned}
\end{equation}
Here $\mathcal{L}_{\mathrm{cons}}$ is the reliability-weighted unlabeled
regression consistency loss, $\mathcal{L}_{\mathrm{cross}}$ compares labeled
examples with unlabeled stable targets, $\mathcal{L}_{\mathrm{weak}}$ and
$\mathcal{L}_{\mathrm{smooth}}$ regularize weak-view predictions, and
$\mathcal{L}_{\mathrm{label}}$ preserves labeled-batch order. The numerical
hyperparameters are listed in the supplement rather than embedded in the
method text.

This formulation keeps ranking as a training signal rather than the primary
metric. The main tables still evaluate regression MAE on held-out test splits.
Ranking enters because candidate ordering can remain informative when absolute
continuous pseudo-labels are uncertain. The component ablation below tests this
claim by removing the anchor, cross-rank, feature-smoothness, and labeled-rank
components from the same DimeNet++ implementation.

\section{Experiments}
We evaluate six scalar materials property tasks: 2D band gap, piezoelectric
tensor, exfoliation energy, elastic anisotropy, experimental formation enthalpy,
and experimental band gap. The CSV benchmark tables were converted from the
processed Matminer task tables released with Chang et al. \cite{chang2022moe};
task-specific primary-source and processed-table provenance are reported in the
supplement. They span electronic, energetic, mechanical, and response-property
regimes and range from 495 to 2,086 labeled structures. Each task is paired with
an MP-5k unlabeled training pool from the Materials Project
\cite{jain2013materials} in the main benchmark; the split sizes are reported in the
supplement. We cross the six tasks with SchNet, DimeNet++, ET, and M3GNet, and
compare supervised training, Pi Model, Mean Teacher, MixMatch, RDA, CLSS, and
\matrank. For every backbone-task block,
checkpoints are selected by validation MAE and then evaluated on the held-out
test split. The main tables therefore report test MAE after validation-only
model selection, with NMAE and method rank used for aggregate comparison.
The five runs use predefined split indices 0--4 with a fixed training random
state; their dispersion therefore measures split sensitivity under a controlled
optimization seed.

\section{Results}
Table~\ref{tab:main-comparison} gives the full held-out test comparison across
six tasks and four backbones. The next tables then test the algorithmic
evidence around \matrank: component ablation on DimeNet++, OOD behavior under
element-level and label-level shifts \cite{koh2021wilds}, and replacement of
MP-5k unlabeled structures with samples from an MP-20-pretrained MatterGen
checkpoint \cite{jain2013materials,zeni2025mattergen}. All reported test values use validation-selected
checkpoints.

The Results section is organized to separate four claims that would otherwise be
collapsed into one aggregate number. The main benchmark asks whether a fixed
\matrank\ objective improves scarce-label regression across encoders and tasks.
The stability view asks whether those gains are accompanied by acceptable
split-level variability. The ablation asks whether the improvement comes from the
combined objective rather than from one removable term. The OOD and generated
pool experiments then test whether the same objective remains usable when either
the evaluation distribution or the unlabeled source changes.

\begin{figure*}[t]
\centering
\includegraphics[width=\textwidth]{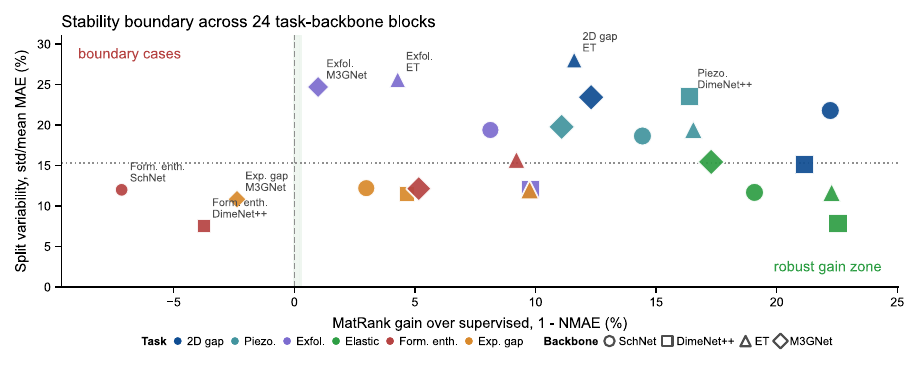}
\caption{Stability boundary for \matrank\ over the 24 task-backbone blocks.
Each point is one task and graph backbone. The horizontal axis reports the
relative gain over supervised training, $1-\mathrm{NMAE}$, and the vertical
axis reports five-split variability as $\mathrm{std}/\mathrm{mean}$ MAE. Colors
indicate materials tasks and marker shapes indicate graph backbones. Dashed
reference lines mark zero gain and the median split variability. This view
separates blocks with stable gains from boundary cases where \matrank\ improves
less or varies more across splits.}
\label{fig:matrank-stability-boundary}
\end{figure*}

\subsection{MatRank Leads the SemiMat Benchmark with Stability Boundaries}
\matrank\ has the lowest overall Avg. NMAE and best average method rank when
the last two columns of Table~\ref{tab:main-comparison} are averaged over the
four backbones. Its average NMAE is $0.896$, compared with $0.949$ for Mean
Teacher, $0.973$ for Pi Model, $0.987$ for MixMatch, $0.988$ for CLSS,
$1.000$ for supervised training, and $1.072$ for RDA. Its average method rank
is $2.208$ across the 24 backbone-task blocks. \matrank\ also has the lowest
backbone-level NMAE for SchNet ($0.901$), DimeNet++ ($0.882$), ET ($0.877$),
and M3GNet ($0.926$), although the best local method remains task-dependent. This
heterogeneity is why the complete matrix is reported rather than only the
aggregate score.

The local pattern is consistent with the benchmark motivation. \matrank\ has the
lowest MAE in many 2D gap, piezoelectric, and elastic-anisotropy blocks, where unlabeled
structure can provide useful geometric regularization. Mean Teacher is the
best-performing non-ranking baseline and remains competitive on formation enthalpy
and experimental gap, where the pointwise teacher signal can be sufficient.
RDA obtains the lowest MAE in selected exfoliation-energy blocks but degrades on formation
enthalpy, which raises its average NMAE. These cases show why a single
aggregate score is not enough for semi-supervised materials regression.

Figure~\ref{fig:matrank-stability-boundary} adds the split-variability axis to
this comparison. Blocks in the lower-right region are the clearest successes:
\matrank\ improves over supervised training while keeping five-split variability
below the median. Blocks near the zero-gain line are treated as boundary cases
rather than hidden successes, even when their aggregate contribution is positive.
This is important for materials tasks because a method that lowers the mean MAE
but substantially increases split sensitivity would be harder to reuse in small
labeled regimes.

\begin{table*}[t]
\centering
\small
\setlength{\tabcolsep}{2.2pt}
\begin{tabular}{lcccccccc}
\toprule
Method & 2D gap & Piezo. & Exfol. & Elastic & Form. enth. & Exp. gap & Avg. NMAE & Avg. rank \\
\midrule
\multicolumn{9}{l}{\textit{SchNet}}\\
Supervised & 0.717$\pm$0.102 & 0.173$\pm$0.022 & 27.612$\pm$3.874 & 0.812$\pm$0.265 & 0.149$\pm$0.018 & 0.346$\pm$0.042 & 1.000 & 5.500 \\
Pi Model & 0.671$\pm$0.066 & 0.172$\pm$0.022 & 27.017$\pm$7.105 & 0.809$\pm$0.208 & \textbf{0.148$\pm$0.017} & 0.345$\pm$0.044 & 0.981 & 3.830 \\
Mean Teacher & 0.666$\pm$0.046 & 0.167$\pm$0.021 & \textbf{24.870$\pm$2.720} & 0.768$\pm$0.259 & 0.148$\pm$0.017 & 0.346$\pm$0.077 & 0.955 & 2.500 \\
MixMatch & 0.681$\pm$0.061 & 0.180$\pm$0.015 & 25.955$\pm$3.793 & 0.791$\pm$0.259 & 0.149$\pm$0.012 & 0.374$\pm$0.064 & 0.997 & 5.000 \\
RDA & 0.684$\pm$0.064 & 0.174$\pm$0.021 & 25.718$\pm$3.397 & 0.815$\pm$0.245 & 0.181$\pm$0.016 & \textbf{0.328$\pm$0.066} & 1.008 & 4.830 \\
CLSS & 0.663$\pm$0.045 & 0.185$\pm$0.015 & 26.488$\pm$4.964 & 0.788$\pm$0.258 & 0.159$\pm$0.021 & 0.339$\pm$0.047 & 0.994 & 4.170 \\
\textbf{MatRank} & \textbf{0.558$\pm$0.121} & \textbf{0.148$\pm$0.028} & 25.367$\pm$4.918 & \textbf{0.657$\pm$0.077} & 0.160$\pm$0.019 & 0.336$\pm$0.041 & \textbf{0.900} & \textbf{2.170} \\
\midrule
\multicolumn{9}{l}{\textit{DimeNet++}}\\
Supervised & 0.639$\pm$0.065 & 0.155$\pm$0.025 & 22.404$\pm$3.209 & 0.798$\pm$0.265 & 0.123$\pm$0.017 & 0.341$\pm$0.032 & 1.000 & 6.000 \\
Pi Model & 0.555$\pm$0.048 & 0.152$\pm$0.018 & 21.599$\pm$4.057 & 0.776$\pm$0.231 & 0.121$\pm$0.018 & 0.325$\pm$0.039 & 0.953 & 3.170 \\
Mean Teacher & 0.566$\pm$0.069 & 0.150$\pm$0.019 & 21.861$\pm$4.271 & 0.770$\pm$0.242 & \textbf{0.118$\pm$0.010} & 0.317$\pm$0.047 & 0.947 & 2.830 \\
MixMatch & 0.592$\pm$0.053 & 0.149$\pm$0.025 & 21.889$\pm$3.465 & 0.767$\pm$0.240 & 0.128$\pm$0.018 & 0.333$\pm$0.048 & 0.973 & 4.000 \\
RDA & 0.611$\pm$0.069 & 0.155$\pm$0.022 & 20.937$\pm$4.899 & 0.776$\pm$0.282 & 0.155$\pm$0.019 & 0.337$\pm$0.064 & 1.018 & 5.000 \\
CLSS & 0.560$\pm$0.033 & 0.160$\pm$0.024 & 22.063$\pm$4.633 & 0.798$\pm$0.221 & 0.126$\pm$0.017 & \textbf{0.316$\pm$0.051} & 0.975 & 4.670 \\
\textbf{MatRank} & \textbf{0.504$\pm$0.076} & \textbf{0.130$\pm$0.030} & \textbf{20.209$\pm$2.430} & \textbf{0.618$\pm$0.048} & 0.128$\pm$0.010 & 0.325$\pm$0.037 & \textbf{0.882} & \textbf{2.330} \\
\midrule
\multicolumn{9}{l}{\textit{ET}}\\
Supervised & 0.722$\pm$0.068 & 0.164$\pm$0.024 & 21.905$\pm$3.970 & 0.863$\pm$0.230 & 0.137$\pm$0.016 & 0.383$\pm$0.086 & 1.000 & 5.670 \\
Pi Model & 0.626$\pm$0.058 & 0.154$\pm$0.018 & 23.018$\pm$5.549 & 0.870$\pm$0.250 & 0.127$\pm$0.013 & 0.381$\pm$0.061 & 0.964 & 4.170 \\
Mean Teacher & \textbf{0.621$\pm$0.035} & 0.167$\pm$0.018 & 20.384$\pm$5.436 & 0.829$\pm$0.218 & \textbf{0.115$\pm$0.017} & 0.370$\pm$0.074 & 0.929 & 2.500 \\
MixMatch & 0.646$\pm$0.064 & 0.160$\pm$0.012 & 21.698$\pm$5.912 & 0.835$\pm$0.242 & 0.129$\pm$0.011 & 0.358$\pm$0.073 & 0.952 & 3.670 \\
RDA & 0.700$\pm$0.110 & 0.173$\pm$0.021 & \textbf{20.259$\pm$3.335} & 0.909$\pm$0.174 & 0.213$\pm$0.011 & 0.427$\pm$0.074 & 1.112 & 5.670 \\
CLSS & 0.668$\pm$0.060 & 0.179$\pm$0.024 & 21.407$\pm$4.028 & 0.859$\pm$0.212 & 0.129$\pm$0.013 & 0.370$\pm$0.084 & 0.983 & 4.500 \\
\textbf{MatRank} & 0.638$\pm$0.179 & \textbf{0.137$\pm$0.026} & 20.967$\pm$5.375 & \textbf{0.671$\pm$0.078} & 0.124$\pm$0.020 & \textbf{0.346$\pm$0.042} & \textbf{0.877} & \textbf{1.830} \\
\midrule
\multicolumn{9}{l}{\textit{M3GNet}}\\
Supervised & 0.732$\pm$0.054 & 0.191$\pm$0.028 & 28.527$\pm$3.119 & 0.825$\pm$0.239 & 0.138$\pm$0.015 & 0.341$\pm$0.059 & 1.000 & 4.170 \\
Pi Model & 0.756$\pm$0.052 & 0.187$\pm$0.030 & 27.855$\pm$2.378 & 0.825$\pm$0.227 & 0.137$\pm$0.006 & 0.340$\pm$0.065 & 0.995 & 3.330 \\
Mean Teacher & 0.703$\pm$0.061 & 0.188$\pm$0.029 & \textbf{27.715$\pm$3.188} & 0.812$\pm$0.227 & \textbf{0.125$\pm$0.007} & \textbf{0.336$\pm$0.058} & 0.965 & \textbf{1.830} \\
MixMatch & 0.806$\pm$0.217 & 0.191$\pm$0.030 & 29.699$\pm$3.619 & 0.832$\pm$0.230 & 0.141$\pm$0.015 & 0.343$\pm$0.059 & 1.028 & 5.830 \\
RDA & 0.737$\pm$0.038 & 0.184$\pm$0.031 & 28.572$\pm$3.761 & 0.852$\pm$0.222 & 0.252$\pm$0.016 & 0.367$\pm$0.065 & 1.149 & 5.670 \\
CLSS & 0.728$\pm$0.033 & 0.191$\pm$0.029 & 28.075$\pm$2.071 & 0.843$\pm$0.233 & 0.140$\pm$0.006 & 0.342$\pm$0.066 & 1.003 & 4.670 \\
\textbf{MatRank} & \textbf{0.642$\pm$0.150} & \textbf{0.170$\pm$0.034} & 28.246$\pm$6.974 & \textbf{0.683$\pm$0.105} & 0.131$\pm$0.016 & 0.349$\pm$0.038 & \textbf{0.926} & 2.500 \\
\bottomrule
\end{tabular}
\caption{Five-split held-out test MAE across four graph backbones. Task columns
report mean$\pm$standard deviation in original units. Avg. NMAE is normalized
by the supervised MAE for the same task and backbone; Avg. rank averages
task-wise method ranks within a backbone. Bold marks the local column best.}
\label{tab:main-comparison}
\end{table*}

\subsection{Reliability and Ranking Components Stabilize MatRank Gains}
Table~\ref{tab:ablation} isolates \matrank\ components on DimeNet++. The full
objective has the best Avg. NMAE, while several ablations remain competitive on
individual tasks. This pattern supports the intended design: the final method
does not depend on one term alone, but on combining anchor reliability,
consistency, cross-set ranking, feature smoothness, and labeled-rank
preservation.

The ablation also clarifies what the method is not claiming. Removing one
component can improve a single task, such as the experimental-gap result for
the label-rank ablation, but it weakens the aggregate behavior. Removing the
anchor pseudo-label, cross-set ranking, feature-smoothness, or labeled-rank term
can improve a local task, but each weakens the aggregate behavior relative to
the full objective. These terms therefore act as stabilizers around a pointwise
regressor rather than as replacements for regression. Avg. rank is recomputed
over the rows retained in this table.

The strongest ablated rows are also informative. The cross-rank and
feature-smoothness ablations remain close to the full objective in average NMAE,
which indicates that \matrank\ is not a brittle sum of unrelated penalties. The
full objective is nevertheless the only row that simultaneously preserves the
best aggregate NMAE and the best average rank. This supports the intended design
choice: reliability weighting protects continuous pseudo-targets, while ranking
terms provide order information when absolute pseudo-label values are uncertain.

\begin{table*}[t]
\centering
\small
\setlength{\tabcolsep}{2.0pt}
\begin{tabular}{lcccccccc}
\toprule
Variant & 2D gap & Piezo. & Exfol. & Elastic & Form. enth. & Exp. gap & Avg. NMAE & Avg. rank \\
\midrule
Supervised & 0.640$\pm$0.065 & 0.155$\pm$0.025 & 22.404$\pm$3.209 & 0.798$\pm$0.265 & \textbf{0.123$\pm$0.017} & 0.341$\pm$0.032 & 1.000 & 6.000 \\
w/o anchor PL & 0.597$\pm$0.153 & 0.125$\pm$0.034 & 21.032$\pm$3.869 & 0.620$\pm$0.036 & 0.133$\pm$0.006 & 0.323$\pm$0.048 & 0.914 & 3.167 \\
w/o cross-rank & 0.543$\pm$0.128 & \textbf{0.122$\pm$0.039} & 20.796$\pm$3.699 & 0.637$\pm$0.061 & 0.127$\pm$0.011 & 0.329$\pm$0.044 & 0.892 & 3.000 \\
w/o feature smooth. & 0.541$\pm$0.085 & 0.128$\pm$0.039 & 21.618$\pm$5.892 & 0.627$\pm$0.076 & 0.131$\pm$0.017 & 0.328$\pm$0.049 & 0.907 & 3.500 \\
w/o label-rank & 0.558$\pm$0.094 & 0.126$\pm$0.040 & 22.106$\pm$6.201 & 0.633$\pm$0.052 & 0.134$\pm$0.009 & \textbf{0.307$\pm$0.057} & 0.909 & 3.833 \\
\textbf{Full MatRank} & \textbf{0.504$\pm$0.076} & 0.130$\pm$0.031 & \textbf{20.209$\pm$2.430} & \textbf{0.618$\pm$0.048} & 0.128$\pm$0.010 & 0.325$\pm$0.037 & \textbf{0.882} & \textbf{2.333} \\
\bottomrule
\end{tabular}
\caption{DimeNet++ component ablation over six tasks. Supervised and Full
MatRank are five-split reference rows; ablated variants report
mean$\pm$standard deviation over three split runs.}
\label{tab:ablation}
\end{table*}

\subsection{MatRank Retains Bounded Gains Under OOD Shifts}
Table~\ref{tab:ood-raw} reports raw held-out test MAE under element-level and
label-level OOD splits. \matrank\ improves all six tasks in both split types,
with larger reductions under the element-level shift and smaller but consistent
reductions under the harder label-tail shift.

The two OOD settings test different stresses. Element-level OOD changes
composition coverage, so gains indicate that the learned representation and
anchor reliability can transfer beyond the element distribution observed during
training. Label-level OOD holds out the high-label tail, a more conservative
stress for any pseudo-label method because unlabeled targets near the tail can
be systematically harder to estimate. The smaller label-level gains are
therefore consistent with the method's bounded claim.

\begin{table*}[t]
\centering
\small
\setlength{\tabcolsep}{2.2pt}
\begin{tabular}{llcccccc}
\toprule
OOD split & Method & 2D gap & Piezo. & Exfol. & Elastic & Form. enth. & Exp. gap \\
\midrule
Element-level & Supervised & 0.743$\pm$0.304 & 0.192$\pm$0.045 & 50.887$\pm$10.677 & 0.696$\pm$0.127 & 0.734$\pm$0.593 & 1.144$\pm$0.510 \\
Element-level & \textbf{MatRank} & \textbf{0.717$\pm$0.277} & \textbf{0.185$\pm$0.046} & \textbf{38.294$\pm$12.944} & \textbf{0.688$\pm$0.152} & \textbf{0.646$\pm$0.614} & \textbf{1.068$\pm$0.474} \\
Label-level & Supervised & 1.877$\pm$0.064 & 0.520$\pm$0.020 & 97.700$\pm$1.291 & 2.995$\pm$0.027 & 0.259$\pm$0.009 & 2.093$\pm$0.191 \\
Label-level & \textbf{MatRank} & \textbf{1.792$\pm$0.038} & \textbf{0.512$\pm$0.015} & \textbf{97.372$\pm$1.104} & \textbf{2.944$\pm$0.041} & \textbf{0.256$\pm$0.010} & \textbf{2.079$\pm$0.147} \\
\bottomrule
\end{tabular}
\caption{DimeNet++ OOD raw test MAE over six tasks. Entries are
mean$\pm$standard deviation in original task units after validation-based
checkpoint selection.}
\label{tab:ood-raw}
\end{table*}

\subsection{Synthetic Crystals Also Benefit Semi-Supervised Regression}
Replacing MP-5k unlabeled structures with samples from an MP-20-pretrained
MatterGen checkpoint \cite{jain2013materials,zeni2025mattergen} does not
collapse \matrank.
As Table~\ref{tab:generated-pool} shows, the generated pool is similar to the
MP-5k pool in aggregate and is slightly better on 2D gap, formation enthalpy,
and experimental gap. The experiment changes only the unlabeled pool; the
algorithm and validation-only checkpoint protocol are unchanged.

This result suggests that \matrank\ is not simply memorizing a particular MP-5k
pool. The generated structures supply a different unlabeled distribution, yet
the same reliability and ranking objective remains usable without retuning.
The effect sizes are small, so the result should be read as evidence of pool
tolerance rather than proof that generated pools are universally better.

Together, these diagnostics make the main comparison more interpretable. The
benchmark table shows the aggregate test outcome, the ablation connects that
outcome to objective components, the OOD table tests shifted evaluation splits,
and the generated-pool table tests a shifted unlabeled source. The evidence
therefore supports \matrank\ as a fixed semi-supervised regression objective
under the current protocol, while keeping the limits of the claim visible.

\begin{table}[t]
\centering
\small
\setlength{\tabcolsep}{2.6pt}
\begin{tabular}{lccc}
\toprule
Task & Supervised & MP-5k & MatterGen \\
\midrule
2D gap & 0.640$\pm$0.065 & 0.504$\pm$0.076 & \textbf{0.504$\pm$0.086} \\
Piezo. & 0.155$\pm$0.025 & \textbf{0.130$\pm$0.031} & 0.131$\pm$0.027 \\
Exfol. & 22.404$\pm$3.209 & \textbf{20.209$\pm$2.430} & 20.304$\pm$2.250 \\
Elastic & 0.798$\pm$0.265 & \textbf{0.618$\pm$0.048} & 0.626$\pm$0.069 \\
Form. enth. & 0.123$\pm$0.017 & 0.128$\pm$0.010 & \textbf{0.123$\pm$0.011} \\
Exp. gap & 0.341$\pm$0.032 & 0.325$\pm$0.037 & \textbf{0.321$\pm$0.034} \\
\bottomrule
\end{tabular}
\caption{DimeNet++ \matrank\ with MP-5k or MP20-pretrained MatterGen-generated
unlabeled structures. Entries are raw test MAE mean$\pm$standard deviation over
five split runs.}
\label{tab:generated-pool}
\end{table}

\section{Discussion}
\method\ and \matrank\ address distinct needs in data-scarce materials
modeling. \method\ makes the comparison controlled across tasks, backbones,
unlabeled pools, splits, checkpoint rules, and aggregation metrics. \matrank\ is
one reproducible algorithm within that benchmark, designed for the failure mode
that continuous pseudo-labels can be numerically precise but unreliable. The
main comparison, component ablation, OOD tests, and generated-pool experiment
together support a bounded claim: one fixed \matrank\ objective gives the best
aggregate held-out test NMAE and method rank across the retained 24
backbone-task blocks.

This claim should be read as evidence for reliability-aware semi-supervised
regression, not as universal dominance. \matrank\ does not have the lowest MAE
in every local task, and ranking is not a substitute for pointwise regression.
Mean Teacher, Pi Model, CLSS, and RDA remain competitive in selected blocks.
The useful finding is narrower: labeled-anchor reliability, weak--strong
consistency, and order-aware losses can be combined without selecting a
different objective for each dataset or backbone.

\paragraph{Implications and limitations.}
The practical implication is that unlabeled crystals should be treated as
structured but uncertain evidence. \matrank\ anchors pseudo-targets to nearby
labeled examples, reduces their weight when local agreement is weak, and uses
relative order when absolute values are less reliable. This design matches
materials screening, where candidate priority can matter even when continuous
property estimates remain noisy. The benchmark design is equally important:
the full task-by-backbone matrix keeps local exceptions visible instead of
turning them into a single favorable summary.

Several choices make the evidence auditable. All primary numbers are held-out
test results from validation-selected checkpoints. The main table keeps
supervised and semi-supervised baselines in the same view. The ablation, OOD,
and generated-pool tables use the same DimeNet++ implementation to ask whether
the components matter, whether the method survives distribution shift, and
whether the unlabeled source can change without retuning. These checks do not
replace larger deployment studies, but they reduce the risk that the aggregate
gain is a reporting artifact.

The same logic motivates how \method\ should be extended. A future method could
improve the encoder, alter the unlabeled pool, change the reliability
estimator, or replace the ranking loss. Those changes should be evaluated as
separate factors rather than folded into one new score. The benchmark contract
therefore makes the reporting unit explicit: task, backbone, seed, split,
unlabeled source, validation rule, and raw test MAE should remain visible
before aggregate NMAE or average rank is interpreted. This discipline is
especially important for heterogeneous materials properties, where an error in
experimental band gap may carry a different screening cost from an error in
exfoliation energy or formation enthalpy.

The OOD and generated-pool studies should be interpreted in the same bounded
way. The OOD results show that the objective remains useful when the evaluation
split changes, but they do not prove invariance to every composition shift or
label-tail regime. The generated-pool results show that \matrank\ can use a
different unlabeled source without algorithm changes, but they do not imply
that generated crystals are always preferable to database crystals. Their role
in the paper is diagnostic: they test whether the main benchmark result
survives two realistic perturbations to the learning setting. This makes the
claim stronger than a single in-distribution table, while keeping it below a
deployment guarantee.

The current scope has clear limits. The tasks are scalar regressions, so
vector, tensor, or distributional targets would need new normalization,
uncertainty reporting, and screening metrics. The generated-pool study tests
one MP20-pretrained MatterGen source and should not be generalized to all
crystal generators or filtering strategies. \matrank\ also cannot certify that
an unlabeled structure is stable, synthesizable, or inside the intended
chemical domain; those checks remain part of data curation and materials
validation. Future benchmark extensions should therefore keep the reporting
unit at the backbone-task-seed level, while separating improvements due to the
encoder, unlabeled pool, semi-supervised objective, and evaluation protocol.

\section{Conclusion}
This paper studies how to learn materials properties from scarce labels and
unlabeled crystals without allowing uncertain pseudo-labels to dominate the
training signal. \method\ provides the controlled benchmark contract, and
\matrank\ provides one reliability-weighted, ranking-aware objective under that
contract. Across six tasks, four graph backbones, and five split runs, the fixed
\matrank\ objective achieves the lowest aggregate held-out test NMAE and best
average method rank under validation-only checkpoint selection. The main
boundary is equally important: the evidence supports \matrank\ as a strong
semi-supervised regression baseline for this benchmark, not as a universal
solution to materials discovery. Future extensions should keep the same
attribution discipline when changing tasks, unlabeled pools, encoders, or
screening metrics.

\end{document}